\documentclass[5p]{elsarticle}

\usepackage{lineno}
\usepackage{hyperref}

\usepackage{microtype}
\usepackage{color}
\usepackage{soul}
\usepackage{subcaption}
\usepackage{amssymb}
\usepackage{makeidx}  
\usepackage{caption}
\usepackage{algorithm, algpseudocode}
\usepackage{amsmath}
\usepackage{multirow}
\usepackage{cleveref}

\algtext*{EndWhile}
\algtext*{EndIf}
\algtext*{EndFor}

\definecolor{dgreen_100}{RGB}{139, 195, 74}
\definecolor{dgreen_90}{RGB}{150, 200, 91}
\definecolor{dgreen_70}{RGB}{173, 212, 127}
\definecolor{dgreen_50}{RGB}{197, 225, 164}
\definecolor{dgreen_25}{RGB}{226, 240, 210}

\newcommand{\inputindent}{\hspace*{\algorithmicindent}\hspace*{\algorithmicindent}\hspace{.16667em} }
\newcommand{\localvarsindent}{\inputindent\inputindent \hspace{.75em}}
\newcommand{\Returns}{\textbf{returns} }

\journal{Pattern Recognition Letters}

\begin{document}
\begin{frontmatter}
\title{\texorpdfstring{$\tau$}{t}-SS3: a text classifier with dynamic n-grams for early risk detection over text streams}
 
\author[1,2]{Sergio G. Burdisso\corref{cor}}
\cortext[cor]{Corresponding author:}
\ead{sburdisso@unsl.edu.ar}
\author[1]{Marcelo Errecalde}
\ead{merreca@unsl.edu.ar}
\author[3]{Manuel Montes-y-G\'omez}
\ead{mmontesg@inaoep.mx}

\address[1]{Universidad Nacional de San Luis (UNSL), Ej\'ercito de Los Andes 950, San Luis, San Lius, C.P. 5700, Argentina}
\address[2]{Consejo Nacional de Investigaciones Cient\'ificas y T\'ecnicas (CONICET), Argentina}
\address[3]{Instituto Nacional de Astrof\'isica, \'Optica y Electr\'onica (INAOE), Luis Enrique Erro No. 1, Sta. Ma. Tonantzintla, Puebla, C.P. 72840, Mexico}

\begin{abstract}
A recently introduced classifier, called SS3, has shown to be well suited to deal with early risk detection (ERD) problems on text streams. It obtained state-of-the-art performance on early depression and anorexia detection on Reddit in the CLEF's eRisk open tasks. SS3 was created to deal with ERD problems naturally since: it supports incremental training and classification over text streams, and it can visually explain its rationale.
However, SS3 processes the input using a bag-of-word model lacking the ability to recognize important word sequences.
This aspect could negatively affect the classification performance and also reduces the descriptiveness of visual explanations.
In the standard document classification field, it is very common to use word n-grams to try to overcome some of these limitations.
Unfortunately, when working with text streams, using n-grams is not trivial since the system must learn and recognize which n-grams are important ``on the fly''. 
This paper introduces $\tau$-SS3, an extension of SS3 that allows it to recognize useful patterns over text streams dynamically. 
We evaluated our model in the eRisk 2017 and 2018 tasks on early depression and anorexia detection.
Experimental results suggest that $\tau$-SS3 is able to improve both current results and the richness of visual explanations.
\end{abstract}
\end{frontmatter}
\section{Introduction}

The analysis of sequential data is a very active research area that addresses problems where data is processed naturally as sequences or can be better modeled that way, such as sentiment analysis,  machine translation, video analytics, speech recognition, and time-series processing.
A scenario that is gaining increasing interest in the classification of sequential data is the one referred to as ``early classification'', in which, the problem is to classify the data stream as early as possible without having a significant loss in terms of accuracy.
The reasons behind this requirement of ``earliness'' could be diverse. It could be necessary because the sequence length is not known in advance (e.g. a social media user's content) or, for example, if savings of some sort (e.g. computational savings) can be obtained by early classifying the input.
However, the most important (and interesting) cases are when the delay in that decision could also have negative or risky implications.
This scenario, known as ``early risk detection'' (ERD) has gained increasing interest in recent years with potential applications in rumor detection~\cite{chen2018rumor, ma2016detecting}, sexual predator detection and aggressive text identification \cite{escalante2017early}, depression detection \cite{losada2017erisk, losada2016test} or terrorism detection \cite{iskandar2017terrorism}.
ERD scenarios are difficult to deal with since models need to support: classifications and/or learning over of sequential data (streams); provide a clear method to decide whether the processed data is enough to classify the input stream (early stopping); and additionally, models should have the ability to explain their rationale since people's lives could be affected by their decisions.

A recently introduced text classifier\cite{burdisso2019}, called SS3, has shown to be well suited to deal with ERD problems on social media streams. Unlike standard classifiers, SS3 was created to naturally deal with ERD problems since: it supports incremental training and classification over text streams and it has the ability to visually explain its rationale. It obtained state-of-the-art performance on early depression, anorexia and self-harm detection on the CLEF eRisk open tasks\cite{burdisso2019, burdisso2019clef}.
However, at its core, SS3 processes each sentence from the input stream using a bag-of-word model. This leads to SS3 lacking the ability to capture important word sequences which could negatively affect the classification performance. Additionally, since single words are less informative than word sequences, this bag-of-word model reduces the descriptiveness of SS3's visual explanations.

The weaknesses of bag-of-words representations are well-known in the standard document classification field, in which word n-grams are usually used to overcome them.
Unfortunately, when dealing with text streams, using word n-grams is not a trivial task since the system has to dynamically identify, create and learn which n-grams are important ``on the fly''.
In this paper, we introduce a variation of SS3, called $\tau$-SS3, which expands its original definition to allow recognizing important word sequences. In \autoref{sec:ss3} the original SS3 definition is briefly introduced. \autoref{sec:tss3} formally introduces $\tau$-SS3, in which the needed equations and algorithms are described.
In \autoref{sec:results} we evaluate our model on the CLEF's eRisk 2017 and 2018 tasks on early depression and anorexia detection.
Finally, \autoref{sec:conclusion} summarizes the main conclusions derived from this work.

\section{The SS3 text classifier}
\label{sec:ss3}

\begin{figure*}[t!]
    \centering
    \includegraphics[width=140mm]{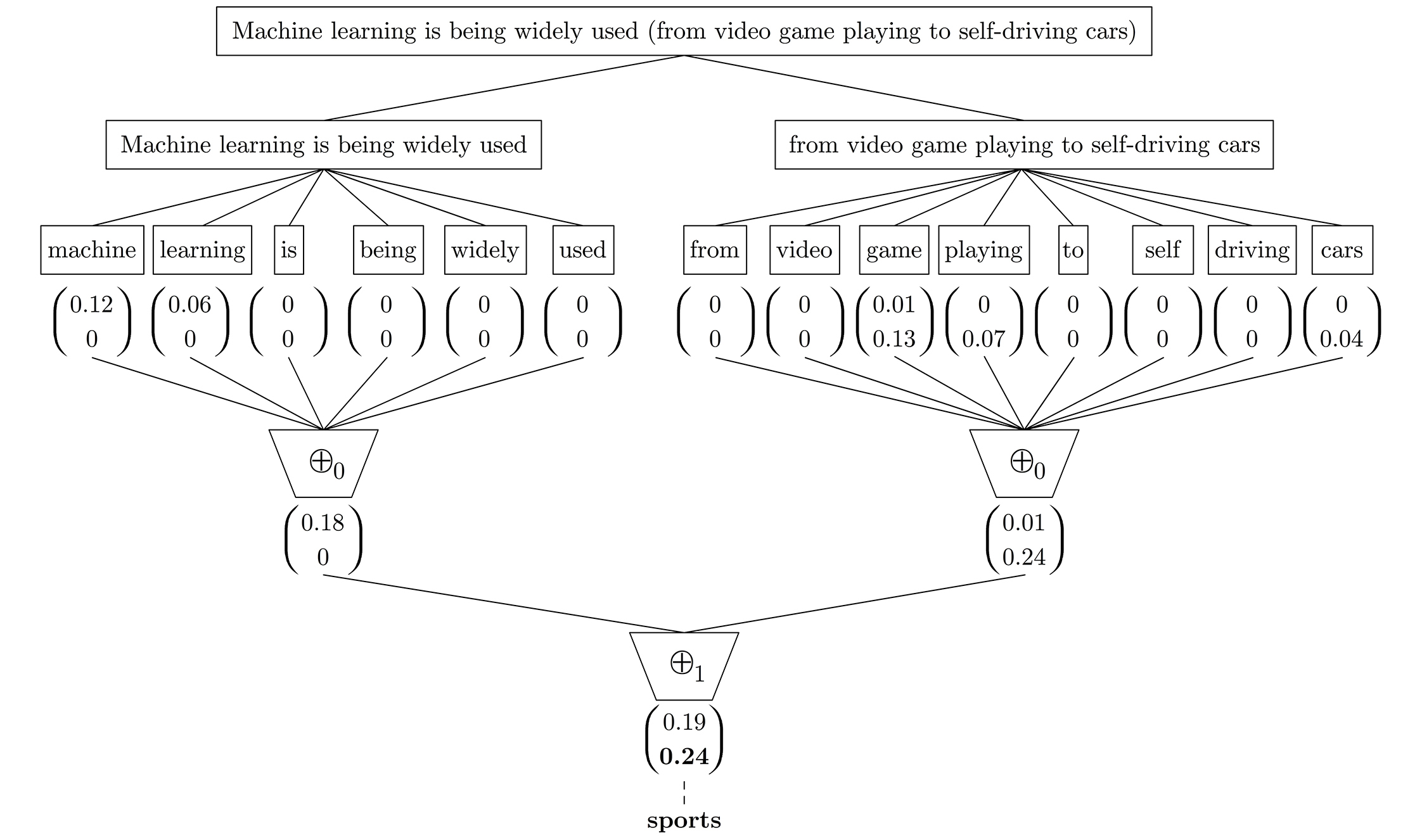}
    \caption{Classification example for categories \emph{technology} and \emph{sports}. In this example, SS3 misclassified the document's topic as $sports$ since it failed to capture important sequences for \emph{technology} like \emph{``machine learning''} or \emph{``video game''}. This was due to each sentence being processed as a bag of words.}
    \label{fig:ss3-example}
\end{figure*}

As it is described in more detail by Burdisso et al. \cite{burdisso2019}, during training and for each given category, SS3 builds a dictionary to store word frequencies using all training documents of the category.
This simple training method allows SS3 to support online learning since when new training documents are added, SS3 simply needs to update the dictionaries using only the content of these new documents, making the training incremental.
Then, using the word frequencies stored in the dictionaries, SS3 computes a value for each word using a function, $gv(w,c)$, to value words in relation to categories. 
$gv$ takes a word $w$ and a category $c$ and outputs a number in the interval [0,1] representing the degree of confidence with which $w$ is believed to \emph{exclusively} belong to $c$, for instance, suppose categories $C= \{food, music, health, sports\}$, we could have:

\

\begin{tabular}{l l} 
$gv($`$sushi$'$, food) = 0.85;$		& $gv($`$the$'$, food) = 0;$\\
$gv($`$sushi$'$, music) = 0.09;$	& $gv($`$the$'$, music) = 0;$\\
$gv($`$sushi$'$, health) = 0.50;$	& $gv($`$the$'$, health) = 0;$\\
$gv($`$sushi$'$, sports) = 0.02;$	& $gv($`$the$'$, sports) = 0;$\\
\end{tabular}

\

Additionally, a vectorial version of $gv$ is defined as: $$\overrightarrow{gv}(w)=(gv(w,c_0), gv(w,c_1), \dots, gv(w,c_k))$$ where $c_i \in C$ (the set of all the categories).
That is, $\overrightarrow{gv}$ is only applied to a word and it outputs a vector in which each component is the word's \emph{gv} for each category $c_i$.
For instance, following the above example, we have:

\

$\overrightarrow{gv}($`$sushi$'$) = (0.85, 0.09, 0.5, 0.02);$

$\overrightarrow{gv}($`$the$'$) = (0, 0, 0, 0);$

\

The vector $\overrightarrow{gv}(w)$ is called the ``\emph{confidence vector} of $w$''.
Note that each category is assigned to a fixed position in $\overrightarrow{gv}$. For instance, in the example above $(0.85, 0.09, 0.5, 0.02)$ is the \emph{confidence vector} of the word ``sushi'' and the first position corresponds to $food$, the second to $music$, and so on.

The computation of $gv$ involves three functions, $lv$, $sg$ and $sn$, as follows:

\begin{equation}
gv(w, c) = lv_\sigma(w, c)\cdot sg_{\lambda}(w, c)\cdot sn_\rho(w, c)
\label{eq:gv}
\end{equation}

\begin{itemize}
\item $lv_\sigma(w, c)$ values a word based on the local frequency of $w$ in $c$. As part of this process, the word distribution curve is smoothed by a factor controlled by the hyperparameter $\sigma$.
\item $sg_{\lambda}(w, c)$ captures the significance of $w$ in $c$. It is a sigmoid function that returns a value close to $1$ when $lv(w, c)$ is significantly greater than $lv(w, c_i)$, for most of the other categories $c_i$; and a value close to $0$ when $lv(w, c_i)$ values are close to each other, for all $c_i$.
The $\lambda$ hyperparameter controls how far $lv(w, c)$ must deviate from the median to be considered significant.
\item $sn_\rho(w, c)$ decreases the global value in relation to the number of categories $w$ is significant to. That is, the more categories $c_i$ to which $sg_{\lambda}(w, c_i)\approx 1$, the smaller the $sn_\rho(w, c)$ value. The $\rho$ hyperparameter controls how severe this sanction is.
\end{itemize}

To keep this paper shorter and simpler we will only introduce here the equation for $lv$ since the computation of both, $sg$ and $sn$, is based only on this function. Nonetheless, for those readers interested in knowing how the $sg$ and $sn$ functions are actually computed, we highly recommend reading the SS3 original paper \cite{burdisso2019}. Thus, $lv$ is defined as:

\begin{equation}
\label{eq:lv-p}
lv_\sigma(w, c) = \bigg(\frac{P(w|c)}{P(w_{max}|c)}\bigg)^\sigma
\end{equation}


Which, after estimating the probability, $P$, by analytical \emph{Maximum Likelihood Estimation}(MLE), leads to the actual definition:


\begin{equation}
\label{eq:lv}
lv_\sigma(w, c) = \bigg(\frac{tf_{w,c}}{max\{tf_c\}}\bigg)^\sigma
\end{equation}

Where $tf_{w,c}$ denotes the frequency of $w$ in $c$, $max\{tf_c\}$ the maximum frequency seen in $c$, and $\sigma \in (0, 1]$ is one of the SS3's hyperparameter.

\

It is worth mentioning that SS3 will learn to automatically ignore stop words since, by definition, $gv(w, c)\approx0$ for all of them.
Therefore, there is no need to manually remove stop words.
Moreover, stop words removal could cause negative effects since in \autoref{eq:lv} we are implicitly evaluating words in terms of stop words.\footnote{That is, words are normalized over the frequency of the most probable one, which will always be a stop word.
Note that the fact that stop words have a similar distribution across all the categories enables us to compute the $gv$ value of a word by comparing its $lv$ value across different categories.}
However, there is nothing in the model preventing us from using any other type of preprocessing, such as stemming, lemmatization, etc.

Finally, during classification, SS3 performs a 2-phase process to classify the input, as it is illustrated in \autoref{fig:ss3-example}.
In the first phase, the input is split into multiple blocks (e.g. paragraphs), then each block is in turn repeatedly divided into smaller units (e.g. sentences, words). Thus, the previously ``flat'' document is transformed into a hierarchy of blocks.
In the second phase, the $\overrightarrow{gv}$ function is applied to each word to obtain the ``level 0'' \emph{confidence vectors}, which then are reduced to ``level 1'' \emph{confidence vectors} by means of a level 0 \emph{summary operator}, $\oplus_0$.\footnote{
Any function $f:2^{\mathbb{R}^n}\mapsto\mathbb{R}^n$ could be used as a \emph{summary operator}, in this example,  vector addition was used.}
This reduction process is recursively propagated up to higher-level blocks, using higher-level \emph{summary operators}, $\oplus_j$, until a single \emph{confidence vector}, $\overrightarrow{d}$, is generated for the whole input.
Finally, the actual classification is performed based on the values of this single \emph{confidence vector}, $\overrightarrow{d}$, using some policy ---for example, selecting the category with the highest \emph{confidence value}.

Note that the classification process is incremental as long as the \emph{summary operator} for the highest level can be computed incrementally.
For instance, suppose that later, a new sentence is appended to the example shown in \autoref{fig:ss3-example}.
Since $\oplus_1$ is the vector addition, instead of processing the whole document again, we could update the already computed vector, $(0.63, 0.07)$, by simply adding the new sentence \emph{confidence vector} to it.
In addition, the \emph{confidence vectors} in the block hierarchy allow SS3 to visually explain the classification if different blocks are painted in relation to their values; this aspect is vital when classification could affect people's lives, humans should be able to inspect the reasons behind the classification.
However, note that SS3 processes individual sentences using a bag-of-word model since the $\oplus_0$ operators reduce the \emph{confidence vectors} of individual words into a single vector. Therefore, SS3 does not take into account any relationship that could exist between individual words, for instance, between ``machine'' and ``learning'' or ``video'' and ``game''. That is, the model cannot capture important word sequences that could improve classification performance, as could have been possible in our example with ``machine learning" or ``self-driving cars''.
In standard document classification scenarios, this type of relationship could be  captured by using variable-length n-grams.
Unfortunately, when working with text streams, using n-grams is not trivial, since the model has to dynamically identify and learn which n-grams are important ``on the fly''. 
In the next section, we will introduce an extension of SS3, called $\tau$-SS3, which is able to achieve it.

\begin{figure*}[t!]
    \centering
    \includegraphics[width=140mm]{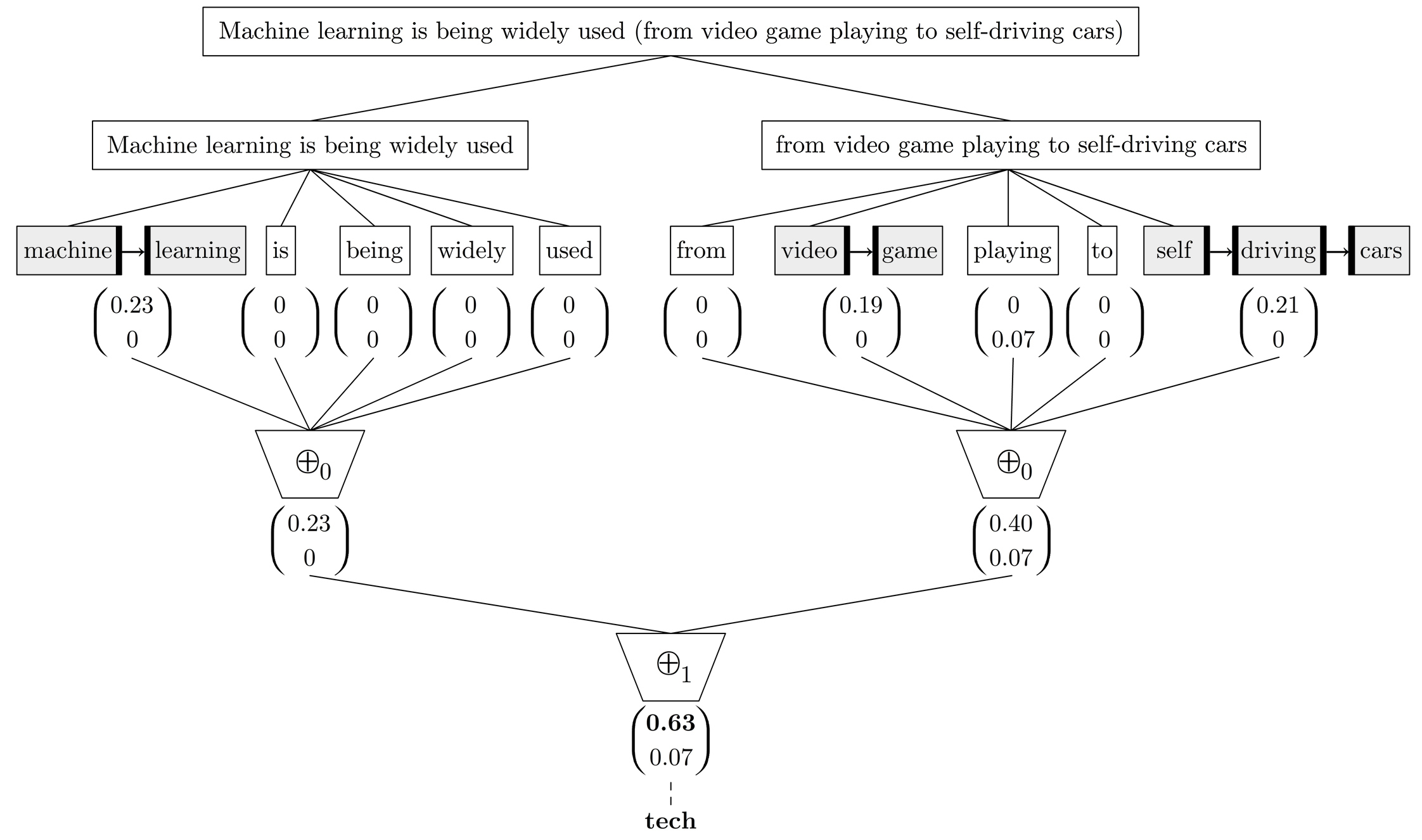}
    \caption{$\tau$-SS3 classification example. Since SS3 now has the ability to capture important word sequences, it is able to correctly classify the document's topic as $tech$.}
    \label{fig:t-ss3-example}
\end{figure*}

\section{The \texorpdfstring{$\tau$}{t}-SS3 text classifier}
\label{sec:tss3}

Regarding the model's formal definition, the only change we need to introduce is a generalized version of the $lv$ function given in \autoref{eq:lv-p}. This is trivial because it only involves allowing $lv$ to value not only words but also sequences of them. That is, in symbols, if $t_k = w_1{\rightarrow}w_2\dots{\rightarrow}w_k$ is a sequence of $k$ words, then $lv$ is now defined as:

\begin{equation}
\label{eq:t-lv-p}
lv_\sigma(t_k, c) = \bigg(\frac{P(w_1w_2\dots w_k|c)}{P(m_1m_2\dots m_k|c)}\bigg)^\sigma
\end{equation}


where $m_1m_2\dots m_k$ is the sequence of $k$ words with the highest probability of occurring given that the category is $c$.


Then, as with \autoref{eq:lv}, the actual definition of $lv$ becomes:

\begin{equation}
lv_\sigma(t_k, c) = \bigg(\frac{tf_{t_k,c}}{max\ tf_{k,c}}\bigg)^\sigma
\label{eq:t-lv}
\end{equation}


Where $tf_{t_k,c}$ denotes the frequency of sequence $t_k$ in $c$ and $max\{tf_{k,c}\}$ the maximum frequency seen in $c$ for sequences of length $k$.


Thus, given any word sequence $t_k$, now we could use the original  \autoref{eq:gv} to compute its $gv(t_k, c)$. For instance, suppose $\tau$-SS3 has learned that the following word sequences have the $gv$ value given below:

\

\begin{tabular}{ll} 
$gv(machine{\rightarrow}learning, tech) = 0.23;$\\
$gv(video{\rightarrow}game, tech) = 0.19;$\\
$gv(self{\rightarrow}driving{\rightarrow}cars, tech) = 0.21;$\\
\end{tabular}

\vspace{5mm}

Then, the previously misclassified example could now be correctly classified, as shown in \autoref{fig:t-ss3-example}.
In the following subsections, we will see how this formal extension is, in fact, implemented in practice. 

\subsection{Training}

\begin{figure*}[t]
    \centering
    \centerline{\includegraphics[width=190mm]{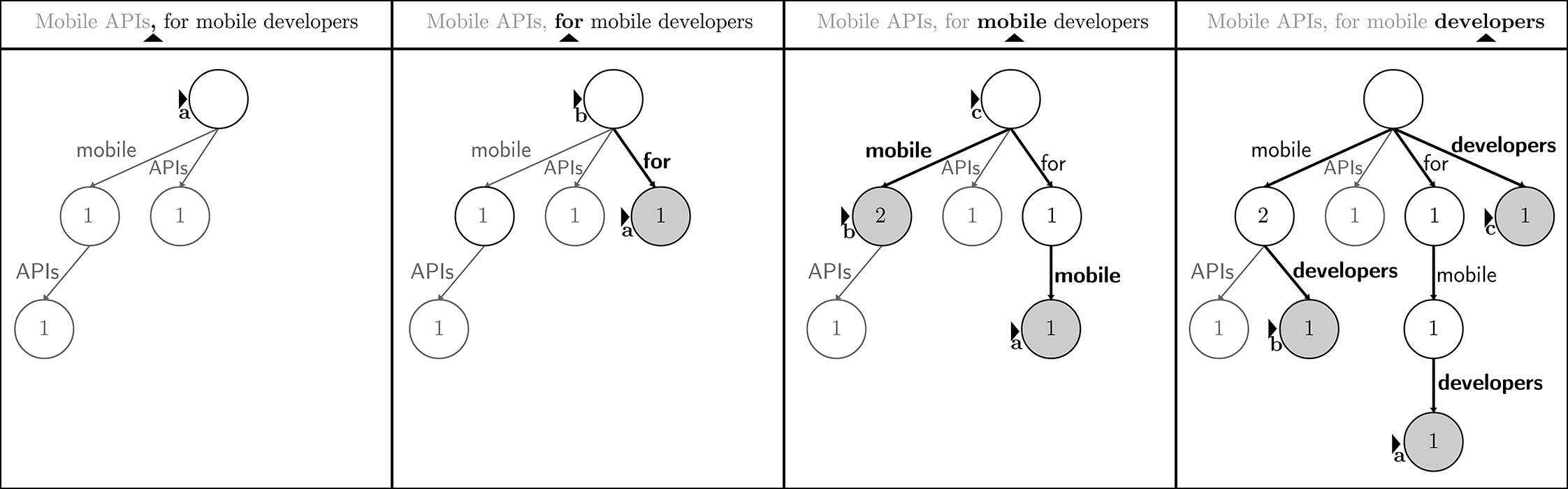}}
    \caption{Training example. Gray color and bold indicate an update. (a) the first two words have been consumed and the tree has 3 nodes, one for each word and one for the bigram ``mobile APIs'', then a comma (,) is found in the input and \autoref{alg:training}'s line 9 and 10 have removed all the cursors and placed a new one, $a$, pointing to the root; (b) the word ``for'' is consumed, a new node for this word is created using the node pointed by cursor $a$ (lines 14), $a$ is updated to point to this new node (line 15 and 20), the next term is read and a new cursor $b$ is created (line 11) in the root; (c) ``mobile'' is consumed, using cursor $b$ the node for this word updated its frequency to 2 (line 16), a new node is created for the bigram ``for mobile'' using cursor $a$, and a new cursor $c$ is created in the root node (line 11); (d) finally, the word ``developers'' is consumed and similarly, new nodes are created for word ``developers'', bigram ``mobile developers'' and trigram ``for mobile developers''.}
    \label{fig:t-ss3-training}
\end{figure*}

\begin{algorithm}[t]
\small
\caption{\small Learning Algorithm. Note that $text$ is a sequence of lexical units (terms) which includes not only words but also punctuation marks.  $MAX\_LVL$ stores the maximum allowed sequence length.}
\label{alg:training}
\begin{algorithmic}[1]
\Statex
\Procedure{Learn-New-Document}{$text$, $category$}
	\State \textbf{input:} $text$, a sequence of lexical units
	\State \inputindent $category$, the category the document belongs to
	\State \textbf{local variables:} $cursors$, a set of prefix tree nodes
	\State
	\State $cursors$ $\gets$ an empty set
	\For{\textbf{each} $term$ \textbf{in} $text$}
		\If{$term$ \textbf{is not} a word}
			\State $cursors$ $\gets$ an empty set
		\Else
		    \State add $category$.\Call{Prefix-Tree}{}.\Call{Root}{} to $cursors$
		    \For{\textbf{each} $node$ \textbf{in} $cursors$}
		        \If{$node$ has \textbf{not} a child for $term$}
		            \State $node$.\Call{Child-Node}{}.\Call{New}{$term$}
		        \EndIf
		        \State $child\_node$ $\gets$ $node$.\Call{Child-Node}{}[$term$]
		        \State $child\_node$.\Call{Freq}{} $\gets$ $child\_node$.\Call{Freq}{} + 1
		        \If{$child\_node$.\Call{Level}{} $\ge$ $MAX\_LVL$}
		            \State remove $node$ from $cursors$
		        \Else
		            \State replace $node$ with $child\_node$ in $cursors$
		        \EndIf
		    \EndFor
		\EndIf
	\EndFor
\EndProcedure
\end{algorithmic}
\end{algorithm}

The original SS3 learning algorithm only needs a dictionary of term-frequency pairs for each category. 
Each dictionary is updated as new documents are processed ---i.e., unseen terms are added and frequencies of already seen terms are updated. Note that these frequencies are the only elements we need to store since to compute $lv(w,c)$ we only need to know $w$'s frequency in $c$, $tf_{w,c}$ (see \autoref{eq:lv}).

Likewise, $\tau$-SS3 learning algorithm only needs to store frequencies of all word sequences seen while processing training documents. More precisely, given a fixed positive integer $n$, it must store information about all word $k$-grams seen during training, with $1 \leq k \leq n$ ---i.e., single words, bigrams, trigrams, etc. To achieve this, the new learning algorithm uses a \emph{prefix tree} (also called \emph{trie})\cite{trie1960, crochemore2009trie} to store all the frequencies, as shown in \autoref{alg:training}. Note that instead of having $k$ different dictionaries, one for each $k$-grams (e.g. one for words, one for bigrams, etc.) we have decided to use a single prefix tree since all n-grams will share common prefix with the shorter ones. Additionally, note that instead of processing the input document $k$ times, again one for each $k$-grams, we have decided to use multiple cursors to be able to simultaneously store all sequences allowing the input to be processed as a stream. Finally, note that lines 8 and 9 of \autoref{alg:training} ensure that we are only taking into account n-grams that make sense, i.e., those composed only of words. All these previous observations, as well as the algorithm intuition, are illustrated with an example in \autoref{fig:t-ss3-training}. This example assumes that the training has just begun for the first time and that the short sentence, \emph{``Mobile APIs, for mobile developers''}, is the first document to be processed. Note that this tree will continue to grow, later, as more documents are processed.

Thus, each category has a \emph{prefix tree} storing information linked to word sequences in which there is a tree's node for each learned k-gram. Note that in \autoref{alg:training}, there will never be more than $MAX\_LVL$ cursors and that the height of the trees will never grow higher than $MAX\_LVL$ since nodes at level 1 store 1-grams, at level 2 store 2-gram, and so on.

Finally, it is worth mentioning that this learning algorithm allows us to keep the original one's virtues. Namely, the training is still incremental (i.e., it supports online learning) since there is no need neither to store all documents nor to re-train from scratch every time new training documents are available, instead, it is only necessary to update the already created trees.

\subsection{Classification}

\begin{algorithm}[t!]
\small
\caption{\small Sentence classification algorithm.
\textproc{Map} applies the $gv$ function to every n-gram in $ngrams$ and returns a list of resultant vectors.
\textproc{Reduce} reduces $ngrams\_cvs$ to a single vector by applying the $\oplus_{0}$ operator cumulatively.}
\label{alg:classification}
\begin{algorithmic}[1]
\Statex
\Function{Classify-Sentence}{$sentence$} \Returns a confidence vector
	\State \textbf{input:} $sentence$, a sequence of lexical units
	\State \textbf{local variables:} $ngrams$, a sequence of n-grams
	\State \localvarsindent $ngrams\_cvs$, confidence vectors
 	\State
	\State $ngrams \gets$ \Call{Parse}{$sentence$}
	\State $ngrams\_cvs \gets$ \Call{Map}{}($gv$, $ngrams$)
	\State \Return \Call{Reduce}{}(\Call{$\oplus_{0}$}{}, $ngrams\_cvs$)
\EndFunction
\Statex \hrulefill
\Function{Parse}{$sentence$} \Returns a sequence of n-grams
    \State \textbf{input:} $sentence$, a sequence of lexical units
    \State \textbf{global variables:} $categories$, the learned categories
    \State \textbf{local variables:} $ngram$, a sequence of words
    \State \localvarsindent $output$, a sequence of n-grams
    \State \localvarsindent $bests$, a list of n-grams
    \State
    \State $cur \gets$ the first term in $sentence$
    \While{$cur$ \textbf{is not} empty}
        \For{\textbf{each} $cat$ \textbf{in} $categories$}
            \State $bests[cat] \gets$ \Call{Best-N-Gram}{$cat$, $cur$}
        \EndFor
        \State $ngram \gets$ the n-gram with the highest $gv$ in $bests$
        \State add $ngram$ to $output$
        \State move $cur$ forward $ngram$.\Call{Length}{} positions
    \EndWhile
    \State \Return $output$
\EndFunction
\Statex \hrulefill
\Function{Best-N-Gram}{$cat$, $term$} \Returns a n-gram
    \State \textbf{input:} $cat$, a category
    \State \inputindent $term$, a cursor pointing to a term in the sentence
    \State \textbf{local variables:} $state$, a node of $cat$.\Call{Prefix-Tree}{}
    \State \localvarsindent $ngram$, a sequence of words
    \State \localvarsindent $best\_ngram$, a sequence of words
    \State
    \State $state \gets$ $cat$.\Call{Prefix-Tree}{}.\Call{Root}{}
    \State add $term$ to $ngram$
    \State $best\_ngram \gets$ $ngram$
    \While{$state$ has a child for $term$}
        \State $state \gets$ $state$.\Call{Child-Node}{}[$term$]
        \State $term \gets$ next word in the sentence
        \State add $term$ to $ngram$
        \If{$gv(ngram, cat) > gv(best\_ngram, cat)$}
            \State $best\_ngram \gets$ $ngram$
        \EndIf
    \EndWhile
    \State \Return $best\_ngram$
\EndFunction
\end{algorithmic}
\end{algorithm}

\begin{figure*}[t!]
    \begin{subfigure}{0.5\textwidth}
        \centering
        \includegraphics[width=90mm]{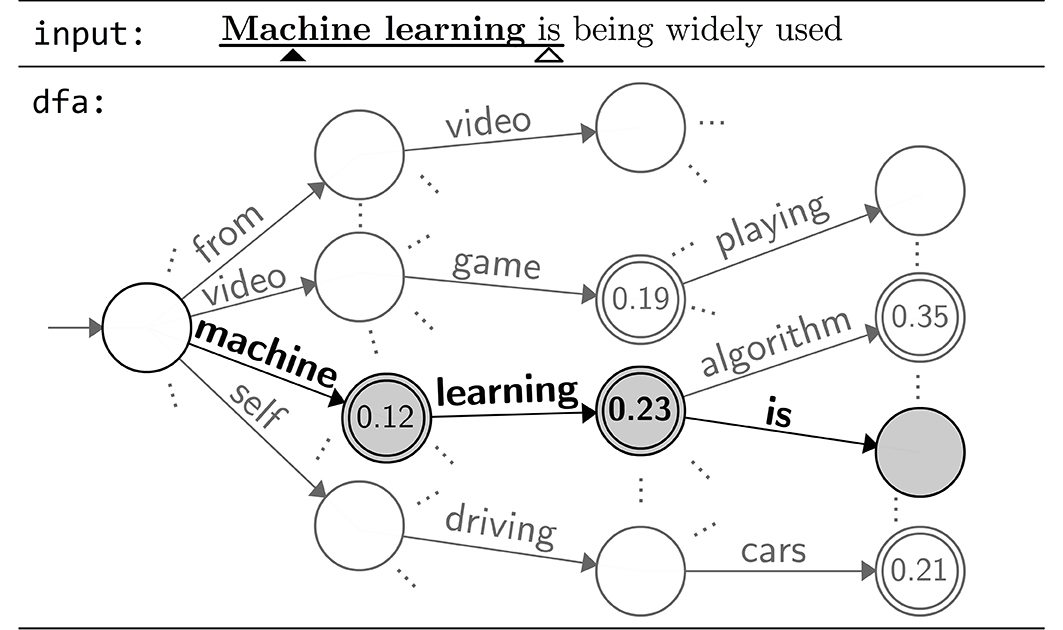}
        \caption{}
        \label{fig:dfa_a}
    \end{subfigure}
    \begin{subfigure}{0.5\textwidth}
        \centering
        \includegraphics[width=90mm]{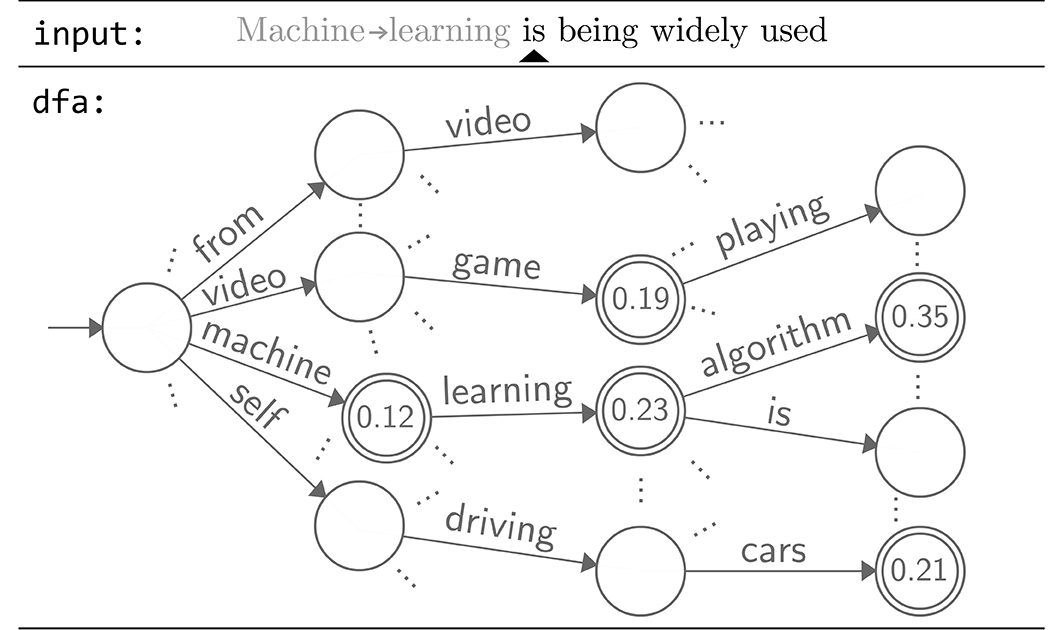}
        \caption{}
        \label{fig:dfa_b}
    \end{subfigure}
\caption{Example of recognizing the best n-gram for the first sentence block of \autoref{fig:t-ss3-example}, ``Machine learning is being widely used''. For simplicity in this example, we only show the  \emph{technology}'s DFA. There are conceptually 2 cursors, the black one ($\blacktriangle$) represents the input cursor and the white one ($\vartriangle$) the ``lookahead'' cursor to feed the automatons. (a) The lookahead cursor has advanced feeding the DFA with 3 words (``machine'', ``learning'', and ``is'') until no more state transitions were available. There were two possible final states, one for ``machine'' and another for ``machine$\rightarrow{}$learning'', the latter is selected since it has the highest $gv$ (0.23); (b) Finally, after the bigram ``machine$\rightarrow{}$learning'' was recognized (see the first two word blocks painted in gray in \autoref{fig:t-ss3-example}), the input cursor advanced 2 positions and is ready to start the process again using ``is'' as the first word to feed the automatons.}
\label{fig:dfa}
\end{figure*}

The original classification algorithm will remain mostly unchanged\footnote{See Algorithm 1 from the original work \cite{burdisso2019}.}, we only need to change the process by which sentences are split into single words, by allowing them to be split into variable-length n-grams.
Also, these n-grams must be ``the best possible ones'', i.e., having the maximum $gv$ value.
To achieve this goal, we will use the prefix tree of each category as a \emph{deterministic finite automaton} (DFA) to recognize the most relevant sequences. Virtually, every node will we considered as a final state if its $gv$ is greater or equal to a small constant $\epsilon$. Thus, every DFA will advance its input cursor until no valid transition could be applied, then the state (node) with the highest $gv$ value will be selected. This process is illustrated in more detail in \autoref{fig:dfa}.

Finally, the formal algorithm is given in \autoref{alg:classification}.\footnote{Note that for this algorithm to be included as part of the SS3's overall classification algorithm, we only need to modify the definition of \textproc{Classify-At-Level}$(text, n)$ defined in Algorithm 1 of the original paper} \cite{burdisso2019} so that, when called with $n \leq 1$, it will call our new function, \textproc{Classify-Sentence}. Note that instead of splitting the sentences into words simply by using a delimiter, now we are calling a \textproc{Parse} function on line 6. \textproc{Parse} intelligently splits the sentence into a list of variable length n-grams. This is done by calling the \textproc{Best-N-Gram} function on line 20 which carries out the process illustrated in \autoref{fig:dfa} to return the best n-gram for a given category.

\section{Experimental results}
\label{sec:results}

\subsection{Tasks and datasets}

Experiments were conducted on three of the CLEF's eRisk open tasks, namely eRisk 2017 and 2018 early depression detection \cite{losada2017erisk, losada2018overview} and eRisk 2018 early anorexia detection \cite{losada2018overview}.
These tasks focused on sequentially processing the content posted by users on Reddit.
Thus, the datasets that were used in these tasks are collections of writings (submissions) posted by a subset of Reddit users (referred to as ``subjects'').
In order to compare the results among different participants, as usual, each dataset is split into a training set and a test set.
Participating research teams were given the training set to train and tune their models offline and were allowed to submit up to five models each.
To carry out the test phase, eRisk organizers divided each user's writing history into 10 chunks.\footnote{Thus, each chunk contained 10\% of the complete user's history.}
Classifiers were given each user's history, one chunk at a time,
and after receiving each chunk, they could either classify the user as depressed/anorexic or wait for the next chunk.

Furthermore, models had to make the correct decision \emph{as early as possible} since their performance was measured taking into account not only the effectiveness but also the delay of their decisions.
Namely, the evaluation metric that was used is called \emph{Early Risk Detection Error} (ERDE).
The ERDE measure was firstly introduced by Losada et al. \cite{losada2017erisk}, it was designed to take into account not only the correctness of decisions but also the delay taken to emit them.
The delay is measured by counting the number ($k$) of different textual items seen before making the binary decision ($d$), which could be positive ($p$) or negative ($n$).
Formally, the ERDE measure is defined as follows:

$$
  ERDE_o(d,k) = \left\{
     \begin{array}{@{}l@{\thinspace}l}
        c_{fp} &\ if\ d=p\ AND\ truth=n\\
        c_{fn} &\ if\ d=n\ AND\ truth=p\\
        lc_o(k)\cdot c_{tp} &\ if\ d=p\ AND\ truth=p\\
        0 &\ if\ d=n\ AND\ truth=n\\
     \end{array}
   \right.
$$

Where the sigmoid \emph{latency cost function}, $lc_o(k)$ is defined by:

$$lc_o(k) = 1 - \frac{1}{1+e^{k - o}}$$

Note that the ERDE measure is parameterized by the $o$ parameter, which acts as the ``deadline'' for decision making, i.e., if a correct positive decision is made in time $k > o$,
it is taken by $ERDE_o$ as if it were incorrect (false positive).
In our case, the performance of all participating models was measured using $ERDE_5$ and $ERDE_{50}$.


\subsection{Implementation details}

The new model implementation was coded in \emph{Python} using only built-in functions and data structures (such as \emph{dict}, \emph{map} and \emph{reduce} functions, etc.). The implementation was added to the official SS3's PyPI package, PySS3\cite{burdisso2019pyss3}, and its source code is available at \href{https://github.com/sergioburdisso/pyss3}{\url{https://github.com/sergioburdisso/pyss3}}.
During experimentation, in order to avoid wasting memory by letting the digital trees grow unnecessarily large, every million words a ``pruning'' procedure was executed in which all the nodes with a frequency less or equal than 10 were removed. We also fixed the maximum n-gram length to 3 (i.e., we set $MAX\_LVL = 3$).\footnote{We tried using different values, from 2 to 10, but the best performance was obtained with $MAX\_LVL = 3$.}

Finally, since we wanted to perform a direct (and fair) comparison against the original SS3 model, we decided to use the same hyperparameter values that were used in the SS3 original paper \cite{burdisso2019}.
Therefore, we set $\lambda=\rho=1$ and $\sigma= 0.455$, which were originally selected by applying a grid search to minimize the $ERDE_{50}$ metric over the training data using 4-fold cross-validation.
Furthermore, as in the original work, vector addition was used as the \emph{summary operator}, $\oplus_j$, for all the levels.
Likewise, the same policy for early classification was also applied, i.e., users were classified as positive as soon as the positive accumulated \emph{confidence value} exceeded the negative one.
Regarding preprocessing, as in the original work, no method was used except for simple accents removal, lowercase conversion, and tokenization.\footnote{
We also tried performing stemming and lemmatization using the Natural Language Toolkit (NLTK), but contrary to what we initially expected, the early classification performance was reduced.
}
This experimental setting ensured that, other than the addition of the variable-length word n-grams, no other factors were influencing the obtained results.



\subsection{Results}

As it is described in more detail in the overview of each task\cite{losada2017erisk, losada2018overview} and the CLEF Working Notes,\footnote{Section ``Early risk prediction on the Internet'' of the CLEF Working Notes for 2017 (\url{http://ceur-ws.org/Vol-1866/}) and 2018 (\url{http://ceur-ws.org/Vol-2125/}).} a total of 180 models were submitted to these three eRisk tasks, ranging from simple to more advanced deep learning models.
For instance, some research groups used simple classifiers such as Multinomial Naive Bayes, Logistic Regression, or Support Vector Machine (SVM) while others made use of more advanced methods such as different types of Recurrent Neural Networks with embeddings, graph-based models, or even ensemble of multiple classifiers.

Results for each one of the three tasks are shown in \autoref{tab:results-d2017}, \autoref{tab:results-d2018}, and \autoref{tab:results-a2018}, respectively.
As can be seen, although not significantly, $\tau$-SS3 improves SS3's performance in all three tasks.
Furthermore, $\tau$-SS3 obtained the best $ERDE_{50}$ values in both depression detection tasks. However, it obtained the second-best value in the anorexia detection task; the best value was obtained by the FHDO-BCSGD\cite{trotzek2018word} model, which consists of a Convolutional Neural Network (CNN) with \emph{fastText} word embeddings.
Regarding $ERDE_{5}$, $\tau$-SS3 also outperformed all the other models in both the anorexia detection and the 2018 depression detection tasks.
However, the best value in the 2018 depression detection task was obtained by the UNSLA\cite{funez2018} model, which consists of an SVM classifier using a novel (time-aware) document representation, called FTVT.
It is worth mentioning that, although not included in the tables, the new model also improved the original SS3's performance in terms of the standard (timeless) measures, precision, recall, and $F_1$.
For instance, in the eRisk 2017 ``Early Depression Detection'' task, $\tau$-SS3's recall, precision and $F_1$ were 0.55, 0.43 and 0.77 respectively, against SS3's 0.52, 0.44 and 0.63.
Additionally, although these values were not the best among all participants, they were quite above the average (0.39, 0.36, and 0.51), which is not bad considering that our hyperparameter values were selected to optimize the $ERDE_{50}$ measure.

\begin{figure}[t!]
	\small
    \begin{subfigure}{90mm}
        \centering
        \includegraphics[width=90mm]{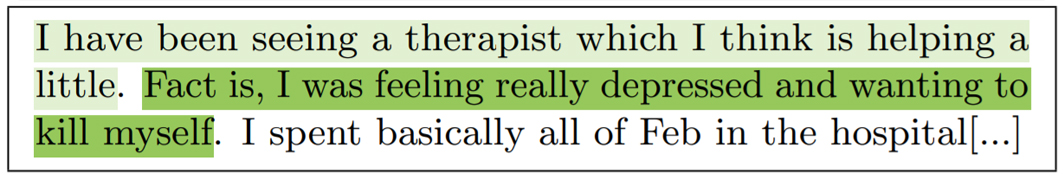}
        \caption{Sentence-level explanation given by SS3}
        \label{fig:subject9579_descriptive_a}
    \end{subfigure}
    \begin{subfigure}{90mm}
        \centering
        \includegraphics[width=90mm]{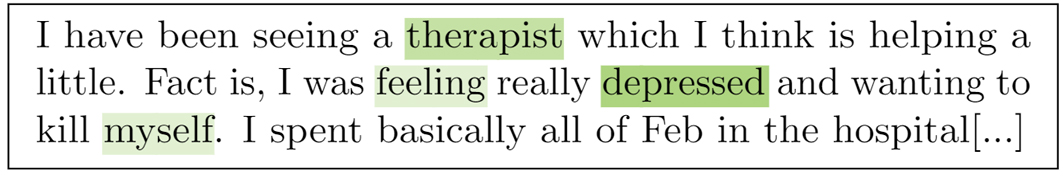}
        \caption{Original word-level explanation, given by $SS3$}
        \label{fig:subject9579_descriptive_b}
    \end{subfigure}
    \begin{subfigure}{90mm}
        \centering
        \includegraphics[width=90mm]{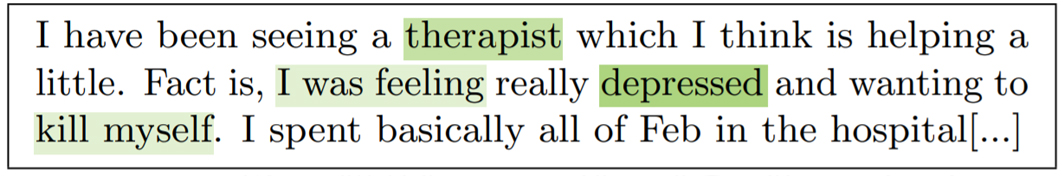}
        \caption{New word-level explanation, given by $\tau$-SS3}
        \label{fig:subject9579_descriptive_c}
    \end{subfigure}
\caption{This figure shows a fragment of the visual explanation given by SS3 in Figure 9 of the original article \cite{burdisso2019}. It shows the subject 9579's writing 60 of the 2017 depression detection task. Blocks are painted proportionally to the true \emph{confidence values} obtained for the ``\emph{depressed}'' category after experimentation. This visual explanation is shown at two different levels: (a) sentences and (b) words. For comparison purposes, in (c), we now show the new visual explanation given by $\tau$-SS3. Note that more useful information is now shown, namely the trigram ``I was feeling'' and the bigram ``kill myself'', improving the richness of visual explanations.}
\label{fig:subject9579_descriptive}
\end{figure}

Results suggest that learned n-grams could contribute to improving the performance of the original model since, although not significantly, $\tau$-SS3 outperformed SS3 in all three tasks. Furthermore, and perhaps more importantly, learned n-grams also contribute to improving visual explanations given by SS3, as illustrated in \autoref{fig:subject9579_descriptive}.\footnote{We have built a live demo to try out $\tau$-SS3 online, available at \href{http://tworld.io/ss3}{\url{http://tworld.io/ss3}}, in which an interactive visual explanation, similar to the one shown in this figure, is given along with the classification result.}

\begin{table}[t]
\centering
\caption{Results on the eRisk 2017 ``Early Depression Detection'' task ordered by ERDE$_{50}$ (the lower, the better). A total of 30 models were submitted by 8 research teams. Here we are only showing the model with best $ERDE_5$ and the model with the best $ERDE_{50}$ of each participating team.}
\label{tab:results-d2017}
\begin{tabular}{l | c c}
\hline
Model & $ERDE_{5}$ & $ERDE_{50}\blacktriangle$ \\ \hline
\textbf{$\tau$-SS3}$\star$ & \textbf{12.6\%} & \textbf{7.70\%}  \\
\textbf{SS3}$\star$ & \textbf{12.6\%} & 8.12\% \\
UNSLA & 13.66\% & 9.68\%  \\
FHDO-BCSGA & 12.82\% & 9.69\% \\
UArizonaD & 14.73\% & 10.23\% \\
FHDO-BCSGB & 12.70\% & 10.39\% \\
UArizonaB & 13.07\% & 11.63\% \\
UQAMD & 13.23\% & 11.98\% \\
GPLC & 14.06\% & 12.14\% \\
CHEPEA & 14.75\% & 12.26\% \\
LyRE & 13.74\% & 13.74\% \\
\hline
\end{tabular}
\end{table}



\begin{table}[t]
\centering
\caption{Results on the eRisk 2018 ``Early Depression Detection'' task ordered by ERDE$_{50}$ (the lower, the better). A total of 44 models were submitted by 11 research teams. Here we are only showing the model with best $ERDE_5$ and the model with the best $ERDE_{50}$ of each participating team.}
\label{tab:results-d2018}
\begin{tabular}{l | c c}
\hline
Model & $ERDE_{5}$ & $ERDE_{50}\blacktriangle$ \\ \hline
\textbf{$\tau$-SS3}$\star$ & 9.48\% & \textbf{6.17\%} \\
\textbf{SS3}$\star$ & 9.54\% & 6.35\%  \\
FHDO-BCSGB & 9.50\% & 6.44\% \\
FHDO-BCSGA & 9.21\% & 6.68\% \\
LIIRB & 10.03\% & 7.09\% \\
PEIMEXC & 10.07\% & 7.35\% \\
UNSLA & \textbf{8.78\%} & 7.39\%  \\
LIIRA & 9.46\% & 7.56\% \\
UQAMA & 10.04\% & 7.85\% \\
LIRMMD & 11.32\% & 8.08\% \\
UDCA & 10.93\% & 8.27\% \\
UPFA & 10.01\% & 8.28\% \\
RKMVERID & 9.97\% & 8.63\% \\
UDCC & 9.47\% & 8.65\% \\
RKMVERIC & 9.81\% & 9.08\% \\
LIRMMA & 10.66\% & 9.16\% \\
TBSA & 10.81\% & 9.22\% \\
TUA1C & 10.86\% & 9.51\% \\
TUA1A & 10.19\% & 9.70\% \\
\hline
\end{tabular}
\end{table}



\begin{table}[t]
\centering
\caption{Results on the eRisk 2018 ``Early Anorexia Detection'' task ordered by ERDE$_{50}$ (the lower, the better). A total of 34 models were submitted by 9 research teams. Here we are only showing the model with best $ERDE_5$ and the model with the best $ERDE_{50}$ of each participating team.}
\label{tab:results-a2018}
\begin{tabular}{l | c c}
\hline
Model & $ERDE_{5}$ & $ERDE_{50}\blacktriangle$ \\ \hline
FHDO-BCSGD & 12.15\% & \textbf{5.96\%}  \\
\textbf{$\tau$-SS3}$\star$ & \textbf{11.31\%} & 6.26\%  \\
FHDO-BCSGE & 11.98\% & 6.61\%  \\
\textbf{SS3}$\star$ & 11.56\% & 6.69\% \\
FHDO-BCSGB & 11.75\% & 6.84\%  \\
PEIMEXB & 12.41\% & 7.79\%  \\
UNSLB & 11.40\% & 7.82\%  \\
RKMVERIA & 12.17\% & 8.63\%  \\
LIIRB & 13.05\% & 10.33\%  \\
LIIRA & 12.78\% & 10.47\%  \\
TBSA & 13.65\% & 11.14\%  \\
UPFA & 13.18\% & 11.34\%  \\
UPFD & 12.93\% & 12.30\%  \\
TUA1C & 13.53\% & 12.57\%  \\
LIRMMB & 14.45\% & 12.62\%  \\
LIRMMA & 13.65\% & 13.04\%  \\
\hline
\end{tabular}
\end{table}

\section{Conclusions and future work}
\label{sec:conclusion}
In this article, we introduced $\tau$-SS3, an extension of the SS3 classification model that allows it to learn and recognize variable-length word n-grams ``on the fly.'' This extension gives $\tau$-SS3 the ability to recognize useful patterns over text streams.
The new model uses a \emph{prefix tree} to store variable-length n-grams seen during training. The same data structure is then used as a DFA to recognize important word sequences as the input is read.
Experimental results showed that, although not significantly, $\tau$-SS3 outperformed SS3 in terms of standard performance metrics as well as the ERDE metrics.
These results suggest that learned n-grams seem to positively contribute to the model's performance as well as to the expressiveness of visual explanations.
Future research should focus on evaluating and analyzing the space and computational complexity of the algorithms and data structures.
Furthermore, it could be interesting to analyze the impact of pruning procedures on both performance and computational resource savings.

\section*{\refname}
\bibliographystyle{splncs04}

\end{document}